\documentclass{article}
\usepackage{spconf,amsmath,graphicx}
\usepackage{amsfonts} 
\usepackage{hyperref}

\title{Conformal Predictions for Human Action Recognition with Vision-Language Models}
\name{Tim Bary*, Cl\'ement Fuchs*, Beno\^it Macq\thanks{*T.~Bary and C.~Fuchs contributed equally. T.~Bary and C.~Fuchs are funded by the MedReSyst project, FEDER and the Walloon Region. Computational resources have been provided by the CÉCI, funded by the F.R.S.-FNRS under Grant No. 2.5020.11 and the Walloon Region.}}
\address{ICTEAM, UCLouvain}
\begin{document}
%
\maketitle
%
%
\begin{abstract}

Human-in-the-Loop (HITL) systems are essential in high-stakes, real-world applications where AI must collaborate with human decision-makers. This work investigates how Conformal Prediction (CP) techniques, which provide rigorous coverage guarantees, can enhance the reliability of state-of-the-art human action recognition (HAR) systems built upon Vision-Language Models (VLMs). We demonstrate that CP can significantly reduce the average number of candidate classes without modifying the underlying VLM. However, these reductions often result in distributions with long tails which can hinder their practical utility. To mitigate this, we propose tuning the temperature of the softmax prediction, without using additional calibration data. This work contributes to ongoing efforts for multi-modal human-AI interaction in dynamic real-world environments.


\end{abstract}
\section{Introduction}
Modern Computer Vision (CV) systems offer high performances over a wide variety of tasks, surpassing human expertise in some cases. However, many applications still rely on Human-In-The-Loop (HITL) frameworks, either to enhance the performance of the underlying CV approach or due to the critical nature of the application, which requires the final decision to be made by a human.
The field of video analysis is no stranger to this dynamic, with HITL frameworks used for video segmentation \cite{oh2019fast}, as well as vehicle identification for autonomous driving \cite{li2023human} or Human Action Recognition (HAR) in the context of video surveillance~\cite{stonebraker2020surveillance}. 

In this context, Conformal Predictions (CP) have garnered significant interest. CP frameworks provide a reduced label set with a robust guarantee on ground truth coverage from the uncertainty estimates of an underlying model, provided a calibration set. This framework has proven beneficial when used in conjunction with human annotators, particularly in object classification tasks \cite{straitouridesigning, cresswell2024conformal}. This makes CP especially useful in critical applications or when a model's performance on a specific task is suboptimal, complementing existing HITL frameworks for tasks such as HAR. 

Historically, the HAR problem was approached with statistical methods relying on features carefully crafted by experts~\cite{dollar2005behavior}. Later on, the advent large-scale datasets as well as high performing deep-learning frameworks lead researcher to use such networks trained in a supervised manner  \cite{wang2016temporal, feichtenhofer2019slowfast, jiang2019stm, arnab2021vivit}. Although the latter showed a significant leap in accuracy, they are largely unable to handle novel classes at test-time. Current state-of-the art approaches \cite{wang2021actionclip} circumvent this problem by relying on extensively pre-trained Vision Language Models (VLMs), which can use textual descriptions of the classes to generate ad-hoc classifiers with strong \textit{zero-shot} performance~\cite{radford_learning_2021}. Recently, foundation models including VLMs have been shown to be strong conformal predictors on general image classification benchmarks \cite{fillioux2024foundation}.

\begin{figure*}[h]
\begin{center}
\includegraphics[width=0.9\textwidth]{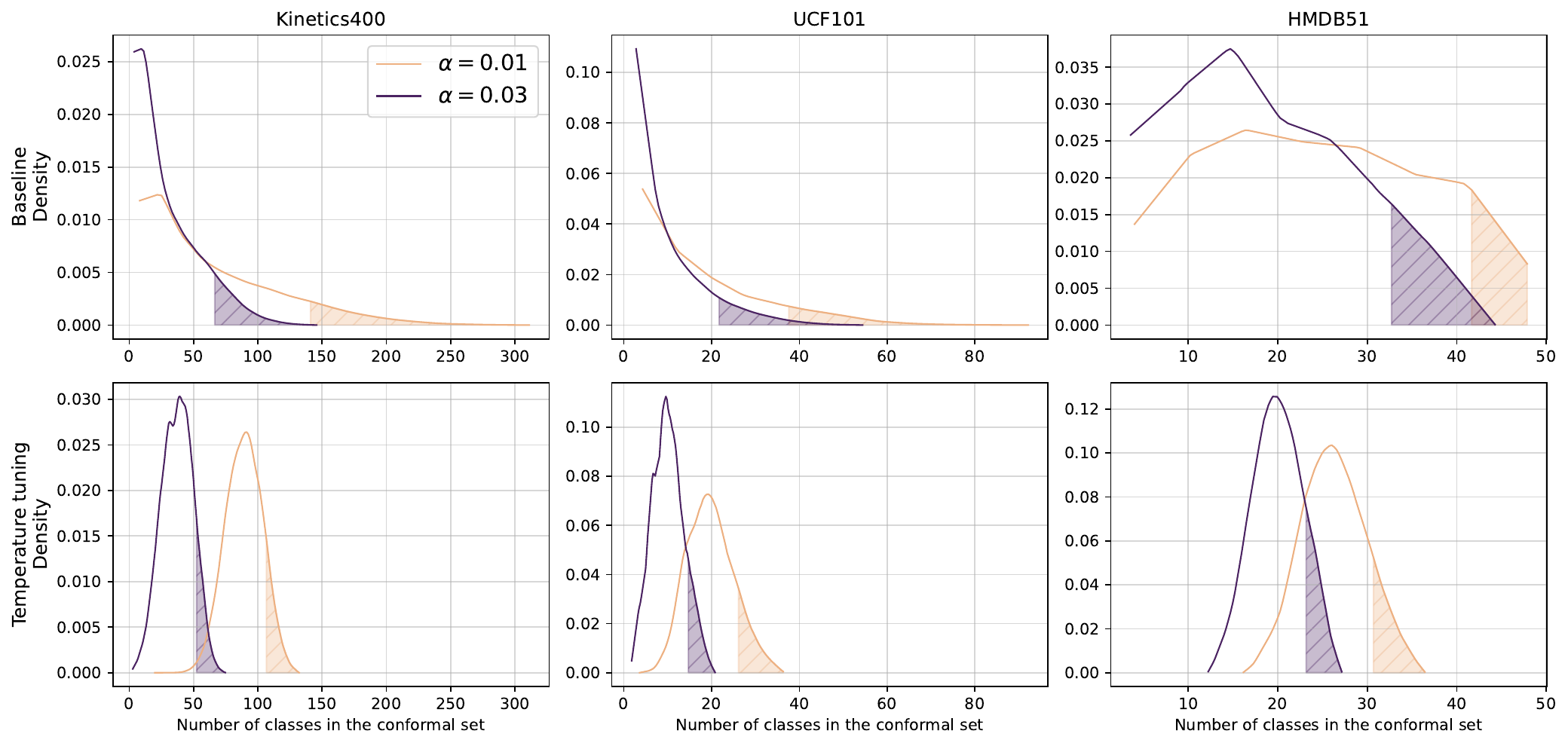}
\end{center}
\vspace{-7mm}
\caption{We show the distribution of conformal sets sizes on three datasets (Kinetics400 with 400 classes, UCF101 with 101 classes and HMDB51 with 51 classes), averaged over 40 folds. The first row shows the distribution of sizes with the baseline, non-tuned temperature parameter. Although the number of classes can be greatly reduced, the distributions are typically long tailed. Comparatively, the second row depicts the results obtained with our temperature tuning strategy. For both rows, the 10\% highest set sizes are shown in the solid color area.}
\label{fig:intro_set_sizes}
\end{figure*}

In this work, we explore the effectiveness of using CPs on top of off-the-shelf VLMs for HAR classification tasks, without any additional fine-tuning. Our results demonstrate that CPs can significantly reduce the number of possible classes for a given video clip, even with high coverage guarantees. We also find that the sizes of the resulting conformal sets typically follow a long-tailed distribution (represented in Figure~\ref{fig:intro_set_sizes}, top). Since human annotation time increases with the number of options available for selection \cite{straitouridesigning, hick1952rate, landauer1985selection}, strategies to shorten this tail and reduce conformal set sizes are valuable, particularly in applications where decision time is constrained, such as live video monitoring. To address this, we highlight the importance of tuning the temperature parameter of the VLM to control the distribution of conformal set sizes. This adjustment can be made using only the calibration set, ensuring no additional data cost for conformal predictor calibration, and preserve the guarantees of the CP framework.

We summarize our contributions as follow: (i) We demonstrate the significant benefits of applying a conformal predictor on off-the-shelf VLMs for human action recognition tasks, and (ii) We propose an efficient, actionable approach to control the long-tail distribution of conformal set sizes. This method, specifically designed for VLMs, preserves the theoretical guarantees of the conformal predictor.

Our code is available at \href{https://github.com/tbary/CP4VLM}{\fontsize{10.8}{10.8}{https://github.com/tbary/CP4VLM}}.

\vspace{-9mm}
\section{Related Work}

\subsection{Conformal Predictions}
Providing reliable confidence estimates for predictions made by deep learning models is essential in many applications. CP offers a solution by producing sets of potential output classes for a given input, with theoretical guarantees—under mild distributional assumptions—about the inclusion of the true class. 

The Least Ambiguous set-valued Classifier~(LAC)~\cite{vovk2005algorithmic,sadinle2019least} computes a non conformity score for each sample $i$ inside a calibration set as $s_i = 1 - y_{i,k^*}$, where $k^*$ is the ground-truth class of sample $i$ and $y_{i,k}$ the $k^{\text{th}}$ component of the soft label $y_i$ within the $K-$simplex $\Delta_K$, as predicted by the model. For a specific error tolerance $\alpha$, the $1 - \alpha$ quantile $\hat{q}$ is derived from the distribution of the non-conformity scores of the calibration samples. The prediction sets for a test sample $i$ is obtained by including all classes $k$ for which $y_{i,k} \geq 1 - \hat{q}$. 

This approach provides theoretical guarantees, specifically that the conformal prediction set for a test sample will, on average, include the true class with a probability of at least $1-\alpha$. This property, known as \textit{coverage}, holds for the dataset as a whole but does not ensure class-specific coverage. For a detailed discussion on the theoretical guarantees and assumptions, refer to~\cite{vovk2005algorithmic,sadinle2019least}. 

The idea of using conformal predictors in the context of HITL systems is not new \cite{zhan2020electronic, cresswell2024conformal}, however, to the best of our knowledge, this work is the first to employ it in conjunction with VLMs for HAR.

\subsection{Human action recognition}
In this work, HAR denotes the problem of classifying video clips or images of a person or group of person performing a well defined action into a set of classes. To address this challenge, early methods focused on the use of handcrafted features \cite{dollar2005behavior, wang2013dense}. Later on, thanks to the emergence of large scale datasets such as Kinetics \cite{carreira2017quo} and advances in machine learning, approaches transitioned to the design of carefully crafted deep-learning architectures trained in a supervised manner. Those architectures include networks based on convolutional building blocks \cite{wang2016temporal, feichtenhofer2019slowfast, jiang2019stm} and more recently transformers~\cite{arnab2021vivit}. 

However, these models are designed to predict a fixed set of classes and do not generalize to settings not seen during training. Current state-of-the-art approaches avoid this problem by relying on VLMs, which can gracefully handle novel classes at test-time due to their extensive pretraining. This paradigm was introduced by \cite{wang2021actionclip}, and later refined by several works such as \cite{ni2022expanding} and \cite{rasheed2023fine}.

\subsection{Vision-Language models}
Vision-language alignment has become a highly influential paradigm for pretraining models that can tackle a broad range of downstream tasks with minimal or no reliance on labeled data. The mapping of textual and visual information into a shared latent space allows for the creation of an ad-hoc classifier by comparing the similarities of input images to encoded textual prompts, the latter being provided by the user. 

Formally, the textual prompts for each of the class $k\in [1,K]$ are processed by the textual encoder to yield the corresponding $\ell_2$ normalized embedding $t_k \in \mathbb{R}^d$, where $d$ is the dimension of the shared latent space for images and texts. The query image $x_i$ is then mapped to its $\ell_2$ normalized embedding $f_i \in \mathbb{R}^d$ through the visual encoder, and similarities scores 
\begin{equation}
l_{i,k} = f_i^\top t_k
\end{equation}
are computed for each class. These logits can be transformed into probabilistic predictions $y_i \in \Delta_K$ with the softmax function as follows:
\begin{equation}
y_{i,k}= \frac{\exp (l_{i,k}/\tau)}{\sum_{j=1}^K \exp (l_{i,j}/\tau)}.
\label{eq:softmax_labels}
\end{equation}

In Equation \ref{eq:softmax_labels}, the temperature parameter $\tau$ controls the sharpness of the resulting probabilistic prediction. When $\tau$ goes to infinity, the soft labels are uniformly equal to $1/K$ while the softmax becomes the $arg max $ operator when $\tau~=~0$. The usual value is $\tau = 0.01$ following~\cite{radford_learning_2021}.

\vspace{-2mm}
\begin{figure}[t]
    \centering
    \includegraphics[width=\linewidth]{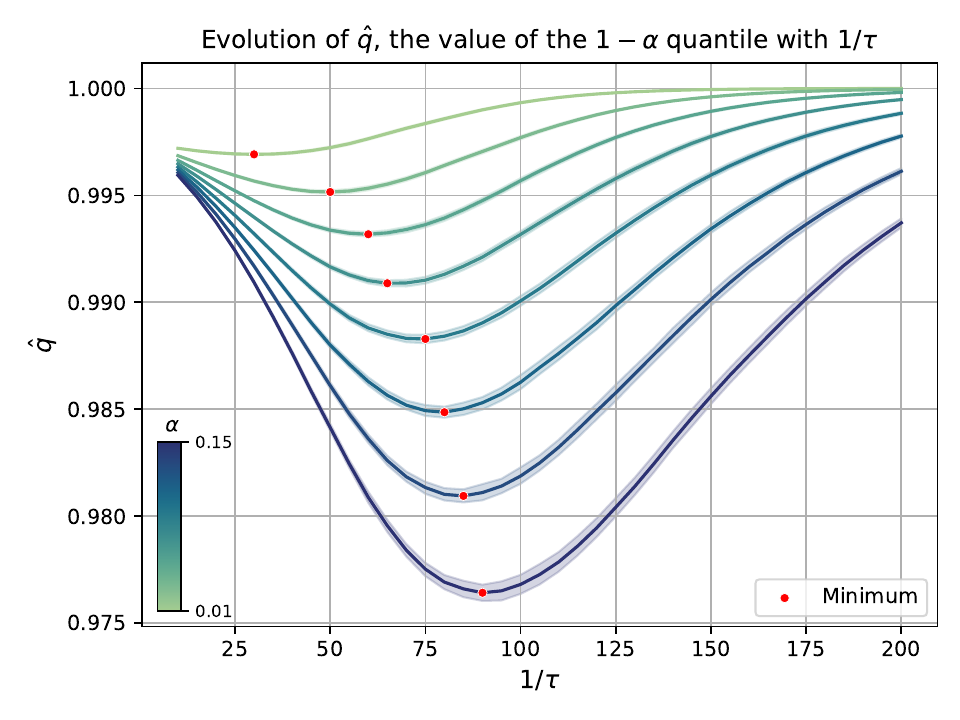}
    \vspace{-9mm}
    \caption{We show the evolution of the $1-\alpha$ quantile of calibration samples conformal scores against the temperature, for varying values of guaranteed error levels $\alpha$. For each value of $\alpha$, the temperature yielding the minimum $1-\alpha$ quantile is selected for the computation of conformal sets at test time.  }
    \label{fig:temp_selection}
    \vspace{-1mm}
\end{figure}
\begin{figure*}[t]
    \centering
    \includegraphics[width=0.92\linewidth]{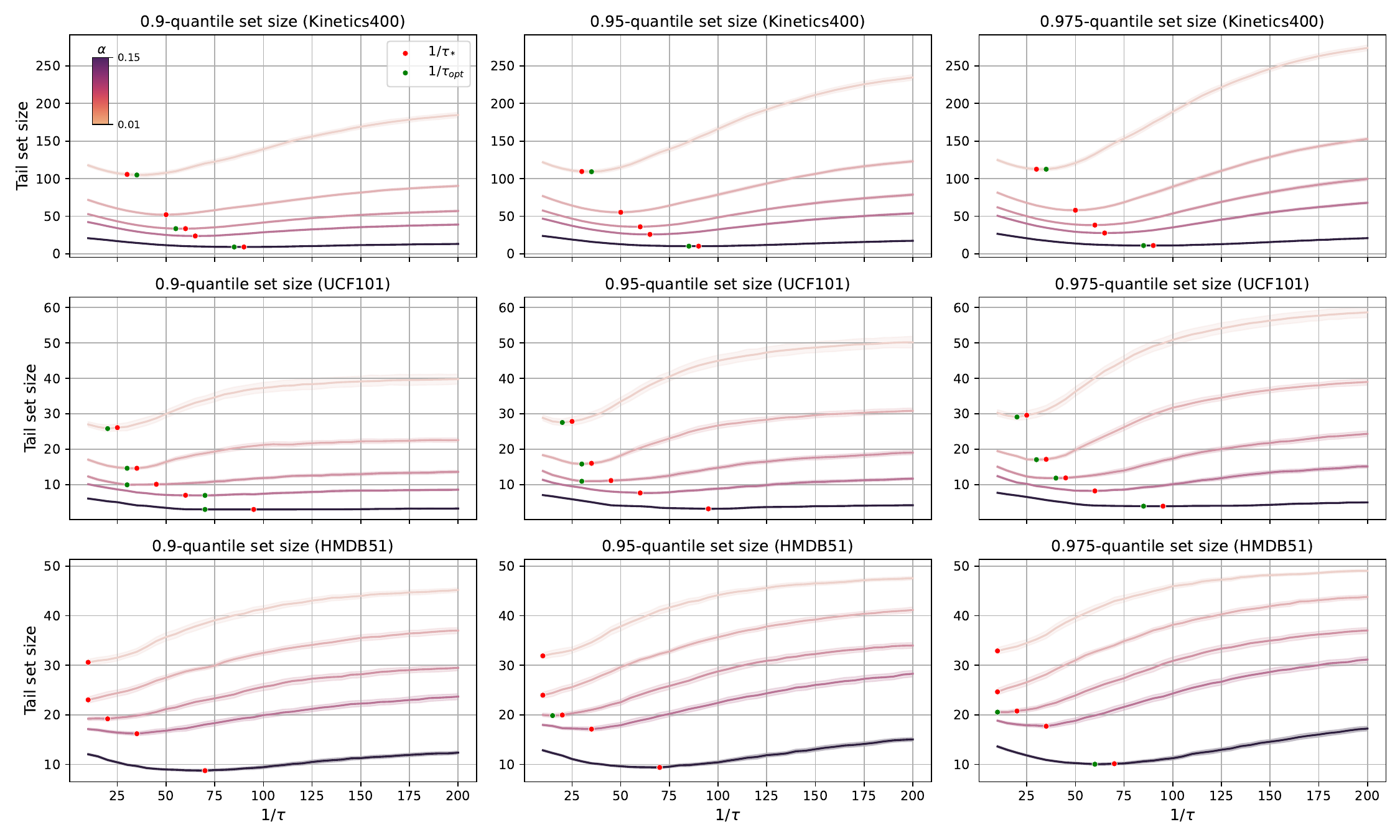}
    \vspace{-4mm}
    
    \caption{We show quantiles for three levels (0.9, 0.95 and 0.975) of the test samples conformal sets sizes distributions as a function of the temperature used to generate soft labels. The value of the quantile corresponding to the temperature estimated following the approach illustrated by Figure \ref{fig:temp_selection} is shown with a red dot, while the value obtained from testing samples is shown with a green dot. When they overlap, only the estimate is shown.}
    \label{fig:tail_size}
\end{figure*}
\begin{figure*}[!h]
    \centering
    \includegraphics[width=0.94\linewidth]{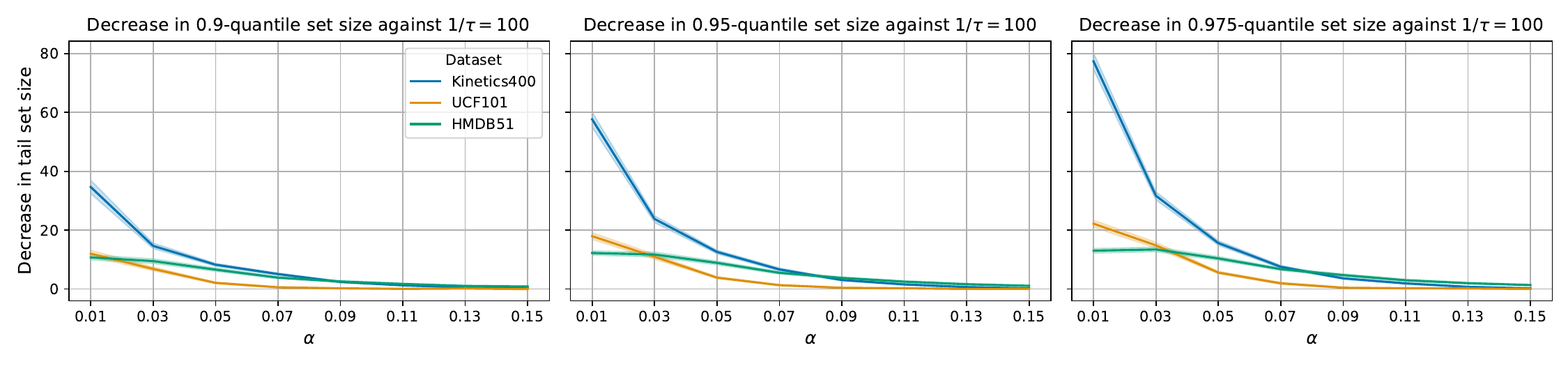}
    \vspace{-5mm}
    
    \caption{We show the decrease of tail set size compared to the value obtained with a non-tuned value of the temperature for three quantile levels (0.9, 0.95, 0.975) for the three datasets used for testing. }
    \label{fig:tail_gain}
    
\end{figure*}
\begin{figure*}[!h]
    \centering
    \includegraphics[width=0.95\linewidth]{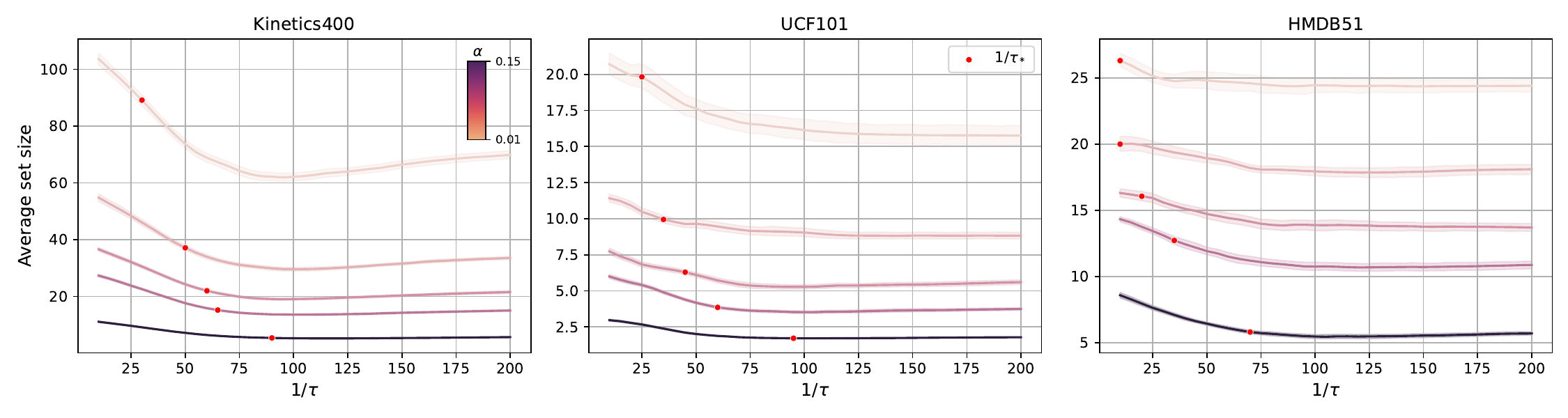}
    \vspace{-5mm}
    \caption{We show the average set size as a function of $1/\tau$ for varying values of $\alpha$. The red dot corresponds to the temperature minimizing the tail set size obtained with the rule illustrated in Figure \ref{fig:temp_selection}. }
    \label{fig:mean_size}
\end{figure*}
\section{Methods}

\subsection{Experimental settings}\label{sec:datasets}
We conduct our experiments using three video clip datasets: HMDB51 (51 classes) \cite{hmdb51}, UCF101 (101 classes) \cite{soomro2012ucf101} and Kinetics400 (400 classes) \cite{carreira2017quo}. For HMDB51 and UCF101, we use all available samples and split them uniformly at random into a 10-shot calibration set and a testing set. For Kinetics400, we apply the same splitting method to the testing and validation samples from the official split from \cite{carreira2017quo}. The results reported are averaged over 40 random seeds for data splitting.

To compute embeddings for each video clip, we extract 10 frames at uniform time intervals, and use the average of their visual embeddings, yielding a single global embedding per video. Note this is the Global Average Pooling (GAP) method described in~\cite{wang2021actionclip}. For text prompts, we use the template: "a photo of a person doing \{class\}." Unless otherwise specified, we use a CLIP model with a ViT-B/16 image encoder. Using this setup, we achieve base accuracies of 54\% on Kinetics400, 69\% on UCF101 and 48\% on HMDB51.

\subsection{Temperature tuning for tail size reduction}
In Figure~\ref{fig:intro_set_sizes}, we observe how a Conformal Predictor, used atop an off-the-shelf VLM significantly reduces the number of classes from which a human expert must choose. However, distribution of set sizes appears long-tailed on the right. This means that, for a small proportion of samples, the number of classes in the conformal set remains relatively large.

Previous research has shown that human decision time increases with the number of available choices \cite{hick1952rate, landauer1985selection}, and many practical scenarios follow \textit{Hick's law}, a logarithmic relationship. More recent studies suggest a sigmoidal model \cite{pavao2016sequence} to capture the dependence of decision time on task uncertainty, for which the number of classes in the conformal sets could serve as a proxy. This highlights the importance of controlling the tail of the set size distribution, especially for applications such as live video monitoring, where the maximum time spent by a human operator to annotate a data point is limited.

As $\tau$, the temperature parameter, affects the predicted soft labels through Equation \ref{eq:softmax_labels}, its variation indirectly influences the distribution of the non-conformity scores derived from the 10-shots calibration sets. Consequently, $\tau$ also impacts the value of $\hat{q}$, the $1-\alpha$ quantile of these non-conformity scores. The relationship between $\hat{q}$ and $\tau$, shown on Figure \ref{fig:temp_selection} for different $\alpha$, is convex. To minimize the tail of the conformal set sizes distribution, we select the temperature $\tau_* = \arg \min_{\tau \in\mathbb{R}} \hat{q}(\tau)$.

\section{Results and Discussion}
\begin{figure}[t]
    \centering
    \includegraphics[width=\linewidth]{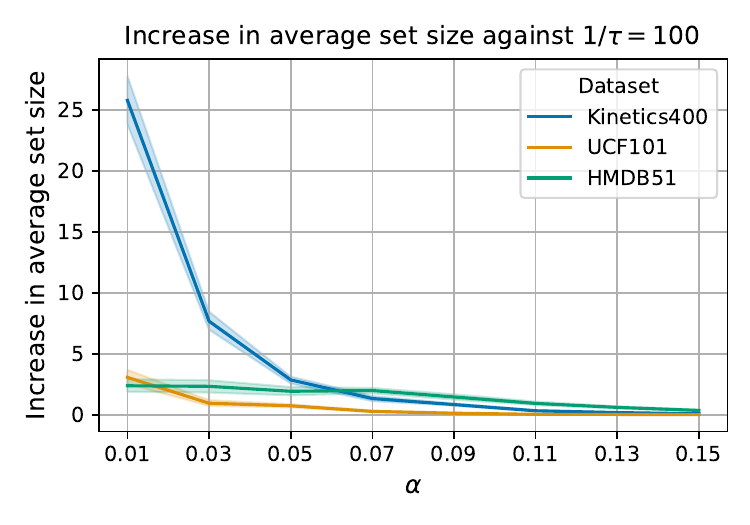}
    \vspace{-10mm}
    \caption{Choosing a temperature that minimizes the tail of conformal sets sizes distribution increases the average set size, illustrating how tuning the temperature is a tradeoff that must take into account the desiredata of the user.}
    \label{fig:mean_loss}
    \vspace{-3mm}
\end{figure}

\begin{figure}[t]
    \centering
    \includegraphics[width=\linewidth]{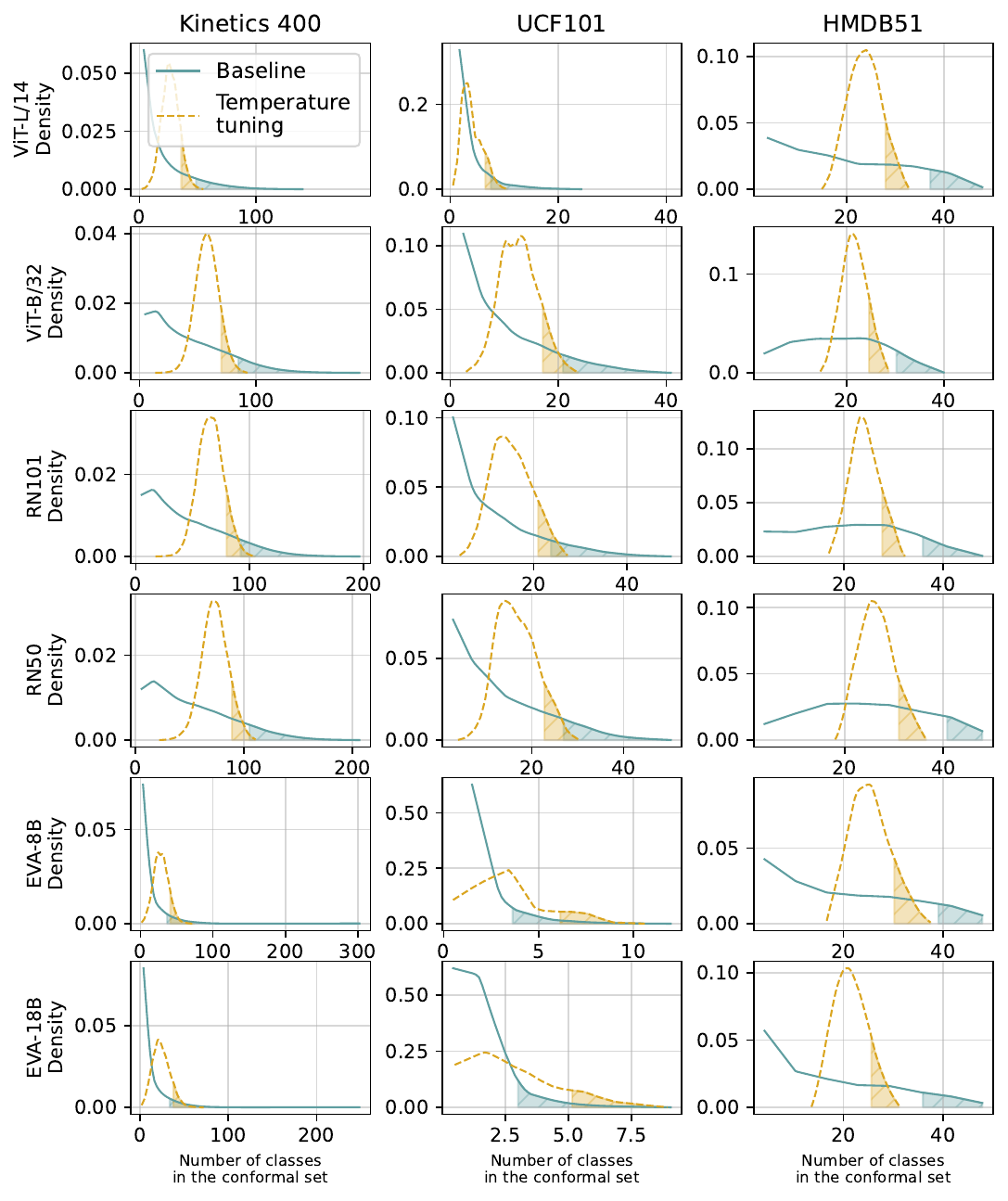}
    \caption{We show that results obtained with other architectures (ViT-L/14, ViT-B/32, ResNet101 and ResNet50), as well as EVA-8B and -18B \cite{sun2024eva} are coherent with the details provided for ViT-B/16. The tuned temperature is obtained according to the rule illustrated in Figure \ref{fig:temp_selection}, for $\alpha = 0.03$. The highest $10\%$ of set sizes are shown with the filled area.}
    \label{fig:other_archs}
    \vspace{-5pt}
\end{figure}
Figure \ref{fig:tail_size} shows how the tail of the conformal set size distribution (quantiles at 0.9, 0.95, and 0.975) varies with the inverse temperature $1/\tau$ across datasets and $\alpha$ values. Red dots mark $1/\tau_*$, our estimate for minimizing tail size, while green dots indicate the true optimum $1/\tau_{opt}$ when it differs. Across all settings, tail sizes at $\tau_*$ were within 0.775 of those at $\tau_{opt}$, validating our method's robustness. Sensitivity to deviations from $1/\tau_{opt}$ increases as $\alpha$ decreases, emphasizing the need for precise tuning at low $\alpha$.

Figure \ref{fig:tail_gain} compares tail size reductions relative to the common $1/\tau=100$ baseline. Gains are largest at smaller $\alpha$, with up to 77 fewer classes for Kinetics400. However, Figure~\ref{fig:mean_size} shows this comes at the cost of larger average set sizes, as $\tau_*$ occurs before the plateau in set size reduction. This trade-off is also reflected in Figure \ref{fig:mean_loss}.

Figure~\ref{fig:intro_set_sizes} illustrates how shifting from $1/\tau=100$ to $1/\tau_*$ affects the distribution in two ways:
(i)~Tail size is minimized at lower $1/\tau$; deviations increase tail size asymmetrically, especially for $1/\tau > 1/\tau_{opt}$, and (ii)~the average set size drops rapidly with $1/\tau$, then levels off.
As optimal values differ for tail and average sizes, a task-specific compromise is needed.

Finally, Figure \ref{fig:other_archs} confirms our method generalizes across CLIP models with both ViT and CNN visual encoders.

\section{Conclusion and Future Works}
In this paper, we demonstrated how integrating Conformal Predictors with off-the-shelf Vision-Language Models (VLMs) can significantly reduce the number of possible classes in Human Action Recognition tasks while maintaining coverage guarantees. Our findings highlight the influence of the temperature parameter $\tau$ on the distribution of conformal set sizes, which mediates a trade-off between low mean, long-tailed distributions and higher mean, shorter-tailed ones.

The relationship between annotation time and the number of classes remains an open question. Classical studies~\cite{hick1952rate, landauer1985selection} suggest a logarithmic dependency on decision time, while alternative models propose a sigmoid relationship with task uncertainty~\cite{pavao2016sequence}. Future research could develop models linking annotation speed and accuracy to conformal set sizes, offering practical insights for optimizing workflows.




\bibliographystyle{IEEEbib}
\bibliography{main}

\begin{thebibliography}{10}

\bibitem{oh2019fast}
Seoung~Wug Oh, Joon-Young Lee, Ning Xu, and Seon~Joo Kim,
\newblock ``Fast user-guided video object segmentation by interaction-and-propagation networks,''
\newblock in {\em Proceedings of the IEEE/CVF Conference on Computer Vision and Pattern Recognition}, 2019, pp. 5247--5256.

\bibitem{li2023human}
Zepeng Li, Dongxiang Zhang, Yanyan Shen, and Gang Chen,
\newblock ``Human-in-the-loop vehicle reid,''
\newblock in {\em Proceedings of the Thirty-Seventh AAAI Conference on Artificial Intelligence and Thirty-Fifth Conference on Innovative Applications of Artificial Intelligence and Thirteenth Symposium on Educational Advances in Artificial Intelligence}, 2023, pp. 6048--6055.

\bibitem{stonebraker2020surveillance}
Michael Stonebraker, Bharat Bhargava, Michael Cafarella, Zachary Collins, Jenna McClellan, Aaron Sipser, Tao Sun, Alina Nesen, K~Solaiman, Ganapathy Mani, et~al.,
\newblock ``Surveillance video querying with a human-in-the-loop,''
\newblock in {\em Proceedings of the Workshop on Human-In-the-Loop Data Analytics with SIGMOD}, 2020.

\bibitem{straitouridesigning}
Eleni Straitouri and Manuel~Gomez Rodriguez,
\newblock ``Designing decision support systems using counterfactual prediction sets,''
\newblock in {\em Forty-first International Conference on Machine Learning}, 2024.

\bibitem{cresswell2024conformal}
Jesse~C Cresswell, Yi~Sui, Bhargava Kumar, and No{\"e}l Vouitsis,
\newblock ``Conformal prediction sets improve human decision making,''
\newblock in {\em Forty-first International Conference on Machine Learning}, 2024.

\bibitem{dollar2005behavior}
Piotr Doll{\'a}r, Vincent Rabaud, Garrison Cottrell, and Serge Belongie,
\newblock ``Behavior recognition via sparse spatio-temporal features,''
\newblock in {\em 2005 IEEE international workshop on visual surveillance and performance evaluation of tracking and surveillance}. IEEE, 2005, pp. 65--72.

\bibitem{wang2016temporal}
Limin Wang, Yuanjun Xiong, Zhe Wang, Yu~Qiao, Dahua Lin, Xiaoou Tang, and Luc Van~Gool,
\newblock ``Temporal segment networks: Towards good practices for deep action recognition,''
\newblock in {\em European conference on computer vision}. Springer, 2016, pp. 20--36.

\bibitem{feichtenhofer2019slowfast}
Christoph Feichtenhofer, Haoqi Fan, Jitendra Malik, and Kaiming He,
\newblock ``Slowfast networks for video recognition,''
\newblock in {\em Proceedings of the IEEE/CVF international conference on computer vision}, 2019, pp. 6202--6211.

\bibitem{jiang2019stm}
Boyuan Jiang, MengMeng Wang, Weihao Gan, Wei Wu, and Junjie Yan,
\newblock ``Stm: Spatiotemporal and motion encoding for action recognition,''
\newblock in {\em Proceedings of the IEEE/CVF international conference on computer vision}, 2019, pp. 2000--2009.

\bibitem{arnab2021vivit}
Anurag Arnab, Mostafa Dehghani, Georg Heigold, Chen Sun, Mario Lu{\v{c}}i{\'c}, and Cordelia Schmid,
\newblock ``Vivit: A video vision transformer,''
\newblock in {\em Proceedings of the IEEE/CVF international conference on computer vision}, 2021, pp. 6836--6846.

\bibitem{wang2021actionclip}
Mengmeng Wang, Jiazheng Xing, and Yong Liu,
\newblock ``Actionclip: A new paradigm for video action recognition,''
\newblock {\em arXiv preprint arXiv:2109.08472}, 2021.

\bibitem{radford_learning_2021}
Alec Radford, Jong~Wook Kim, Chris Hallacy, Aditya Ramesh, Gabriel Goh, Sandhini Agarwal, Girish Sastry, Amanda Askell, Pamela Mishkin, Jack Clark, et~al.,
\newblock ``Learning transferable visual models from natural language supervision,''
\newblock in {\em International conference on machine learning}. PMLR, 2021, pp. 8748--8763.

\bibitem{fillioux2024foundation}
Leo Fillioux, Julio Silva-Rodr{\'\i}guez, Ismail~Ben Ayed, Paul-Henry Courn{\`e}de, Maria Vakalopoulou, Stergios Christodoulidis, and Jose Dolz,
\newblock ``Are foundation models for computer vision good conformal predictors?,''
\newblock {\em arXiv preprint arXiv:2412.06082}, 2024.

\bibitem{hick1952rate}
William~E Hick,
\newblock ``On the rate of gain of information,''
\newblock {\em Quarterly Journal of experimental psychology}, vol. 4, no. 1, pp. 11--26, 1952.

\bibitem{landauer1985selection}
Thomas~K Landauer and Daniel~W Nachbar,
\newblock ``Selection from alphabetic and numeric menu trees using a touch screen: breadth, depth, and width,''
\newblock {\em ACM SIGCHI Bulletin}, vol. 16, no. 4, pp. 73--78, 1985.

\bibitem{vovk2005algorithmic}
Vladimir Vovk, Alexander Gammerman, and Glenn Shafer,
\newblock {\em Algorithmic learning in a random world}, vol.~29,
\newblock Springer, 2005.

\bibitem{sadinle2019least}
Mauricio Sadinle, Jing Lei, and Larry Wasserman,
\newblock ``Least ambiguous set-valued classifiers with bounded error levels,''
\newblock {\em Journal of the American Statistical Association}, vol. 114, no. 525, pp. 223--234, 2019.

\bibitem{zhan2020electronic}
Xianghao Zhan, Zhan Wang, Meng Yang, Zhiyuan Luo, You Wang, and Guang Li,
\newblock ``An electronic nose-based assistive diagnostic prototype for lung cancer detection with conformal prediction,''
\newblock {\em Measurement}, vol. 158, pp. 107588, 2020.

\bibitem{wang2013dense}
Heng Wang, Alexander Kl{\"a}ser, Cordelia Schmid, and Cheng-Lin Liu,
\newblock ``Dense trajectories and motion boundary descriptors for action recognition,''
\newblock {\em International journal of computer vision}, vol. 103, pp. 60--79, 2013.

\bibitem{carreira2017quo}
Joao Carreira and Andrew Zisserman,
\newblock ``Quo vadis, action recognition? a new model and the kinetics dataset,''
\newblock in {\em proceedings of the IEEE Conference on Computer Vision and Pattern Recognition}, 2017, pp. 6299--6308.

\bibitem{ni2022expanding}
Bolin Ni, Houwen Peng, Minghao Chen, Songyang Zhang, Gaofeng Meng, Jianlong Fu, Shiming Xiang, and Haibin Ling,
\newblock ``Expanding language-image pretrained models for general video recognition,''
\newblock in {\em European Conference on Computer Vision}. Springer, 2022, pp. 1--18.

\bibitem{rasheed2023fine}
Hanoona Rasheed, Muhammad~Uzair Khattak, Muhammad Maaz, Salman Khan, and Fahad~Shahbaz Khan,
\newblock ``Fine-tuned clip models are efficient video learners,''
\newblock in {\em Proceedings of the IEEE/CVF Conference on Computer Vision and Pattern Recognition}, 2023, pp. 6545--6554.

\bibitem{hmdb51}
H.~Kuehne, H.~Jhuang, E.~Garrote, T.~Poggio, and T.~Serre,
\newblock ``Hmdb: A large video database for human motion recognition,''
\newblock in {\em 2011 International Conference on Computer Vision}, 2011, pp. 2556--2563.

\bibitem{soomro2012ucf101}
K~Soomro,
\newblock ``Ucf101: A dataset of 101 human actions classes from videos in the wild,''
\newblock {\em arXiv preprint arXiv:1212.0402}, 2012.

\bibitem{pavao2016sequence}
Rodrigo Pav{\~a}o, Joice~P Savietto, Jo{\~a}o~R Sato, Gilberto~F Xavier, and Andr{\'e}~F Helene,
\newblock ``On sequence learning models: Open-loop control not strictly guided by hick’s law,''
\newblock {\em Scientific reports}, vol. 6, no. 1, pp. 23018, 2016.

\bibitem{sun2024eva}
Quan Sun, Jinsheng Wang, Qiying Yu, Yufeng Cui, Fan Zhang, Xiaosong Zhang, and Xinlong Wang,
\newblock ``Eva-clip-18b: Scaling clip to 18 billion parameters,''
\newblock {\em arXiv preprint arXiv:2402.04252}, 2024.

\end{thebibliography}

\end{document}